\documentclass[conference,compsoc]{IEEEtran}
\usepackage{xcolor}
\usepackage{hyperref}

\ifCLASSOPTIONcompsoc
  \usepackage[nocompress]{cite}
\else
  \usepackage{cite}
\fi
\usepackage{graphicx}
\usepackage{booktabs}
\usepackage{enumitem}
\ifCLASSINFOpdf
\else
\fi
\begin{document}
%

\title{Accuracy comparison across face recognition algorithms: \\Where are we on measuring race bias?}



%
\author{\IEEEauthorblockN{Jacqueline G. Cavazos\IEEEauthorrefmark{1},
P. Jonathon Phillips\IEEEauthorrefmark{2},
Carlos D. Castillo\IEEEauthorrefmark{3} and 
Alice J. O'Toole\IEEEauthorrefmark{1}}
\IEEEauthorblockA{\IEEEauthorrefmark{1}School of Behavioral and Brain Sciences\\
The University of Texas at Dallas,
Richardson, TX\\ Email: jacqueline.cavazos@utdallas.edu}
\IEEEauthorblockA{\IEEEauthorrefmark{2}National Institute of Standards and Technology, 
Gaithersburg, MD}
\IEEEauthorblockA{\IEEEauthorrefmark{3}University of Maryland Institute for Advance Computer Studies\\
College Park, MD}}


\maketitle

\begin{abstract}

Previous generations of face recognition algorithms differ in accuracy for images of different races (race bias). Here, we present the possible underlying factors (data-driven and scenario modeling) and methodological considerations for assessing race bias in algorithms. We discuss data-driven factors (e.g., image quality, image population statistics, and algorithm architecture), and scenario modeling factors that consider the role of the ``user'' of the algorithm (e.g., threshold decisions and demographic constraints). To illustrate how these issues apply, we present data from four face recognition algorithms (a previous-generation algorithm and three deep convolutional neural networks, DCNNs) for East Asian and Caucasian faces. First, dataset difficulty affected both overall recognition accuracy and race bias, such that race bias increased with item difficulty. Second, for all four algorithms, the degree of bias varied depending on the identification decision threshold. To achieve equal false accept rates (FARs), East Asian faces required higher identification thresholds than Caucasian faces, for all algorithms. Third, demographic constraints on the formulation of the distributions used in the test, impacted estimates of algorithm accuracy. We conclude that race bias needs to be measured for individual applications and we provide a checklist for measuring this bias in face recognition algorithms.
\end{abstract}


%
\IEEEpeerreviewmaketitle

\section{Introduction}
\vskip -0.25cm
Scientists have over 50 years of experience studying the effects of race on human face recognition ability. People recognize faces of their ``own'' race more accurately than faces of ``other'' races \cite{malpass1969recognition}. This phenomenon is referred to as the ``other-race effect'' (ORE). The study of race bias in computational algorithms, likewise, has a nearly 30-year history that converges on the following finding. Nearly all of the face recognition algorithms studied over the past 30 years show some performance differences as a function of the race of the face. As race bias is investigated in deep convolutional neural network (DCNN) algorithms, it is important to consider the lessons learned from both human- and machine-based studies of race bias in face recognition over the last many decades \cite{malpass1969recognition,o1991simulating}. 

First, we review the combined effects of subject and race of face on face recognition by humans. Second, we review race bias findings for face recognition algorithms. These findings can guide us as we move from previous-generation (simpler) algorithms to present-day DCNNs. These latter algorithms require massive amounts of training data and employ large numbers of local, non-linear computations. Third, we discuss critical, but often overlooked, considerations in measuring face recognition bias in algorithms. Fourth, we will apply the lessons we have learned from the past to measure performance bias in DCNN-based face recognition systems. Specifically, we present novel data from three DCNNs and one previous-generation algorithm, using a dataset in which the challenge level of the comparison items varies. This allows us to measure the effects of two races on performance, as a function of item difficulty. This analysis is especially important for newer DCNN algorithms, which show high accuracy for all but the most challenging stimulus items. We will see that concerns about race bias are magnified, as item difficultly increases. We conclude with a checklist and discussion of considerations to bear in mind when assessing bias in algorithms. Our goal is to better understand how to measure and interpret race bias in face recognition algorithms.   

\subsection{Other-race effect - Humans}
\vskip -0.25cm
The ORE for humans has been found across multiple racial/ethnic groups using different methodological paradigms \cite{malpass1969recognition,meissner2001thirty}. This effect is measured in an experimental paradigm, whereby subjects of different races are tested on their ability to recognize faces of two (or more) races. Formally, the effect is defined by a statistical interaction between the race of the subject and the race of the face. This interaction shows a relative accuracy advantage for own-race faces over other-race faces.

Previous research has demonstrated that expertise in recognizing faces comes, in part, from meaningful experiences with the faces across the lifespan \cite{brigham1985role,bukach2012individuation}, beginning in infancy \cite{kelly2005three,kelly2007other}. Indeed, race biases in face identification have been found across different age groups \cite{pezdek2003children,sangrigoli2005reversibility,yi2016children}. Thus, psychologists have concluded that the power to recognize faces with high accuracy comes from experience discriminating among individuals from a homogeneous population of highly similar faces (i.e., faces of one's own race) \cite{meissner2001thirty,chiroro1995investigation,bukach2012individuation,anzures2014own,tham2017other}. This experience enables the development of neural features that maximize encoding differences among faces within the group. By this account, the ORE for humans is due to the fact that we represent faces of our own-race with a highly effective specialized set of features. These features are not well suited to encode the unique character of other-race faces. Consequently, the experience-dependent nature of human expertise for faces is both a strength and weakness of the human perceptual system. Experience helps us to tune our perceptual systems to rely on features that are optimally suited to faces of our own race, but at the cost of identification performance for faces of other races.

{\bf Lesson 1.} The fact that the ORE has been replicated across multiple races of subjects and faces, combined with findings that experience affects face recognition accuracy in predictable ways, leads us to two conclusions. First, human face recognition ability is characterized by an ORE, whereby performance is relatively more accurate for faces of one's own race. Second, psychological findings support the assumption that faces of all races should be equally recognizable, if one applies appropriate features to analyze a particular race of faces.

\subsection{Racial Bias - Algorithms}
\vskip -0.25cm
Given that humans show an other-race effect for face recognition, it is not surprising that previous research has investigated the effects of race on the accuracy of face recognition algorithms \cite{furl2002face,givens2004features,beveridge2008focus,phillips2011other,klare2012face,el2016face,S_2019_CVPR_Workshops,grother2019face}. For obvious reasons, these effects focus generally on race bias (accuracy differences as function of stimulus race) and not the other-race effect (interaction between the race of the face and race of stimulus). To clarify, as we use it here, {\it race bias} denotes algorithm accuracy differences across groups of faces that vary in race. The integration of state-of-the-art face recognition algorithms in applied settings (e.g., airports) underscores the importance of measuring the effects of race on the accuracy of face recognition algorithms. In what follows, we divide this review of race bias in face recognition algorithms into studies that examine previous-generation algorithms and those that study DCNNs (cf., also for brief overview of this topic see \cite{cavazosIPstrategies}).  

\subsubsection{Pre-DCNN Algorithms}
Differences in the accuracy of face recognition algorithms as a function of race have been reported consistently since the early 1990's. One of the first studies to demonstrate a race bias in algorithms examined early neural network models based on auto-associative learning and principal components analysis (PCA) \cite{o1991simulating}. This study showed that experience-based computational models are influenced by race, but only when the basic features used to encode faces were derived from the statistical structure of the training data. The model was trained with either Asian faces, as the minority race, and Caucasian faces, as the majority race, or vice-versa. The authors found greater identification accuracy for majority-race faces than for minority-race faces, regardless of whether Caucasian or Asian faces were the majority/minority. This suggests that identification accuracy was greater for the race with which the model had greater ``experience''.  

A decade later, researchers examined the performance of face recognition models from the early 2000's for faces of different races \cite{furl2002face}. These models likewise showed differential accuracy as a function of the race of the face, but only when the features used to represent faces were derived from training. Models based on dynamic link architectures, which use preset (hard-coded) features from the visual image, did not show bias, whereas models based on PCA did \cite{WiskottFellousKrugerMalsburg97}. 

Race bias for algorithms has been studied also using race as a covariate \cite{givens2004features,beveridge2008focus}. In 2004, Givens et al. found that covariates, such as race, differentially affected algorithm accuracy  \cite{givens2004features}. Covariates were measured for three algorithms \cite{turk1991face,moghaddam1997probabilistic,wechsler2012face}. Non-Caucasian subjects were easier to recognize than Caucasian subjects. Similarly, in a subsequent study, the three best algorithms from a 2004-2006 algorithm competition \cite{beveridge2009factors} showed that images of non-Caucasians (with the exception of African-Americans\footnote{Terms used to describe racial/ethnic groups are used as they appear in the original published papers.}) were recognized more accurately than Caucasian images. Comparable results were found using a fused algorithm of the top three performing algorithms in a 2006 algorithm competition \cite{beveridge2008focus}. Here, non-Caucasian faces (mostly Asian) were identified more accurately than Caucasian faces. However, the effect of race was smaller than the effect of other factors, such as face size and environment (indoor or outdoor setting). Taken together, these studies show that race, as a covariate, impacts algorithms in ways that are not easy to predict.

In 2011, Phillips et al. further explored the effects of race and found that the origin of the algorithm (i.e., the part of the world where it was developed) mediates race bias in algorithms \cite{phillips2011other}. They compared two fused algorithms: a Western algorithm (created from a fusion of 8 Western algorithms) and an East Asian algorithm (created from a fusion of 5 East Asian algorithms). Both algorithms demonstrated an other-race effect for face identification. The Western algorithm performed more accurately for Caucasian faces and the East Asian algorithm was more accurate for East Asian faces. This is the only study that shows that algorithm origin can predict race bias. The algorithms tested in this study were ``black-box'' algorithms (the algorithms and their implementations were unknown to the researchers who tested for race bias). Differences in the racial composition of the training datasets for Western and 
East Asian algorithms may have contributed to the race bias. 

Perhaps the best known and most comprehensive comparison done on pre-DCNN algorithms was reported in 2012 \cite{klare2012face}. In that study, Klare et al., examined the role of demographics on the accuracy of three commercial off-the-shelf algorithms (COTS), in addition to three in-house algorithms (two non-trainable and one trainable algorithm). Researchers tested effects of race (Black, White, and Hispanic), gender (male/female) and age (young: 18 to 30, middle-aged: 30 to 50, and old:  50 to 70). The results converged on the finding that performance for young, Black, and female faces suffered relative to other demographic groups across all algorithms. Additionally, the authors found that equitable training across all groups (for the trainable algorithms) reduced the effects of specific demographics biases, but did not eliminate them. 

\subsubsection{DCNN Algorithms}

In 2014, the accuracy and generalizability of face recognition algorithms increased markedly with the introduction of algorithms based on DCNNs. These networks employ a series of pooling and convolution operations across multiple layers of simulated neurons. The result of the computations is a compressed representation of a face at the top-layer of network \cite{krizhevsky2012imagenet}. This face descriptor can be examined directly to evaluate the quality of face codes as a function of race and other demographic variables.

Given the prominence of automatic face recognition in social media and security, it is important to know whether these new algorithms show race biases similar to those seen in previous-generation algorithms. To date, only a few studies have examined race bias in DCNNs trained for face recognition \cite{el2016face,cook2019demographic,S_2019_CVPR_Workshops,howard2019effect,grother2019face}. In 2016, El Khiyari et al. evaluated algorithm accuracy using single- and multi-class demographic groups including: gender (male/female), age (young, middle age, older adult), and race (Caucasian/Black). Two algorithms were tested: a COTS face recognition algorithm and a publicly available DCNN (VGG-Face algorithm \cite{parkhi2015deep}) \cite{el2016face}. In the single-class demographics group, accuracy for both algorithms was lower for female, Black, and young groups. These results replicated previous research on older generation algorithms \cite{klare2012face}. Notably, although VGG-Face had a greater overall verification accuracy, it also performed more accurately for images of Caucasians than for images of Black individuals, and showed more race bias than the COTS algorithm. For multi-class demographic group comparisons, accuracy varied widely. Over all comparisons, however, accuracy was greatest for middle-aged White males and was lowest for young Black females.

Cook et al. (2019) tested the effect of demographics on eleven commercial systems (specific algorithms not disclosed) \cite{cook2019demographic}. Each algorithm acquired images from 363 subjects (mostly Black or African-American and White individuals), across multiple demographic groups. Covariate results showed that skin reflectance had the greatest impact on performance. Lower (darker) skin reflectance was associated with longer acquisition times and lower same-identity distribution similarity scores across all systems. Following this, Howard et al. (2019) tested Black or African-American and White faces on a ``leading commercial algorithm''\cite{howard2019effect}. False accept rates (FARs) were greater for White males than for Black males. These results are at odds with previous and subsequent literature (see \cite{klare2012face,el2016face,S_2019_CVPR_Workshops,grother2019face}). The authors suggest that these contradictory results could be explained by the diverse age range in their dataset.

Krishnapriya et al. \cite{S_2019_CVPR_Workshops} compared accuracy for African American and Caucasian faces across four algorithms: two COTS algorithms, VGG-Face (a DCNN), and a ResNet-based DCNN \cite{he2016deep}. The newer of the two COTS and the ResNet algorithm performed best on African American faces. By contrast, VGG-Face and the older COTS performed better on Caucasian faces. Further evidence of race bias was found in the {\it thresholds functions} for FARs, which differed for African American faces and Caucasian faces. (Threshold functions will be discussed in the next section). In the same study, the effect of photo quality on estimates of race bias was also examined \cite{S_2019_CVPR_Workshops}. Results showed that overall accuracy improved when only International Civil Aviation Organization (ICAO)-complaint images were used in the analysis. These results provide additional evidence of race bias in newer DCNNs, as well as a first look at how image quality can affect accuracy performance. 

Most recently, a comprehensive report of demographic effects on face recognition algorithms was released by the National Institute of Standards and Technology (NIST) \cite{grother2019face}. State-of-the-art algorithms from industry and academics volunteered to be tested. Performance was measured on four United States (U.S.) government face image databases that consisted of: (a) U.S. domestic mugshots, (b) immigration benefits application photographs from a global population, (c) Visa application photographs, and (d) border crossing photographs of travelers entering the U.S. At the time of release, the authors reported results ``on over 18.27 million images of 8.49 million people from 189 (mostly commercial) algorithms from 99 developers'' (pp.1)\footnote{As of this writing, the NIST test is on-going, and accepting new algorithm submissions for evaluation, see \url{https://face.nist.gov}.}.

The mugshot database was labeled with ethnic/racial metadata. For the other three data sets, the majority race of the country-of-origin of the photo served as the proxy for the demographic label. The authors limited the countries in their analysis to those countries where a single race of ethnicity dominated. The report lists experimental results on all four data sets, their demographic labels, and numerous scenarios. Across these experimental conditions, there is one common meta-result, race bias over the algorithms tested varies substantially, sometimes over orders of magnitude between the least and most biased algorithm in an experiment. This meta-result strongly recommends that system users measure bias for each task. Because of the variability in bias, the detailed information NIST provides on the performance of these face recognition algorithms is highly valuable for researchers, algorithm developers, users, and policy makers.

{\bf Lesson 2.} Nearly all face recognition algorithms tested in the past 30 years show performances differences as a function of the race of the faces tested. In some cases, algorithms have shown a ``human-like'' interaction between the geographic origin of the algorithm and face race.

\subsection{Measuring Face Identification Accuracy}
\vskip -0.25cm
Before we consider the measurement of race bias, we digress briefly to discuss the standard procedures used for measuring the accuracy of face recognition algorithms, {\it in general}. As we will see, standard methods for measuring algorithm accuracy partly underlie the difficulties we have in estimating race bias in these systems. See \cite{bowyer2019face} for further discussion on this topic.

The standard approach to measuring the accuracy of face recognition algorithms is based on Signal Detection Theory (SDT) \cite{Swets1964-SWESDA}. The problem is formulated as a face-verification task whereby pairs of images, either of the same person or of two different people, are compared. The algorithm generates a similarity score that serves to index the likelihood that the images show the same person (high similarity) or two different people (low similarity). Accuracy is measured based on the degree of overlap between the similarity score distributions for {\it different-identity} pairs and for {\it same-identity} pairs (See Fig 1).

Algorithm accuracy is summarized by the receiving operating characteristic (ROC) curves and synopsized as the area under this curve (AUC). A lower AUC score indicates a larger overlap between the two distributions, which suggests poorer discriminability. A higher AUC score indicates a less overlap between the two distributions, and greater discriminability. AUC = 0.5 indicates chance performance. AUC = 1.0 indicates perfect accuracy. ROC and AUC scores provide robust measures of {\it overall} algorithm accuracy over the entire population of face image pairs in the distributions.

In applications, identification decisions require a {\it threshold similarity score}, over which an image pair is judged as an identity ``match''. Once a threshold is set, additional measures of identification accuracy become relevant. These measures include false rejection rate (FRR) (proportion of same-identity pairs that have been misjudged as different-identity pairs) and false accept rate (FAR) (proportion of different-identity pairs that have been misjudged as same-identity pairs). Threshold placement determines algorithm accuracy at specific FRR and FAR (see Fig. 1). 

\begin{figure}
\centering
\includegraphics[width = \linewidth]{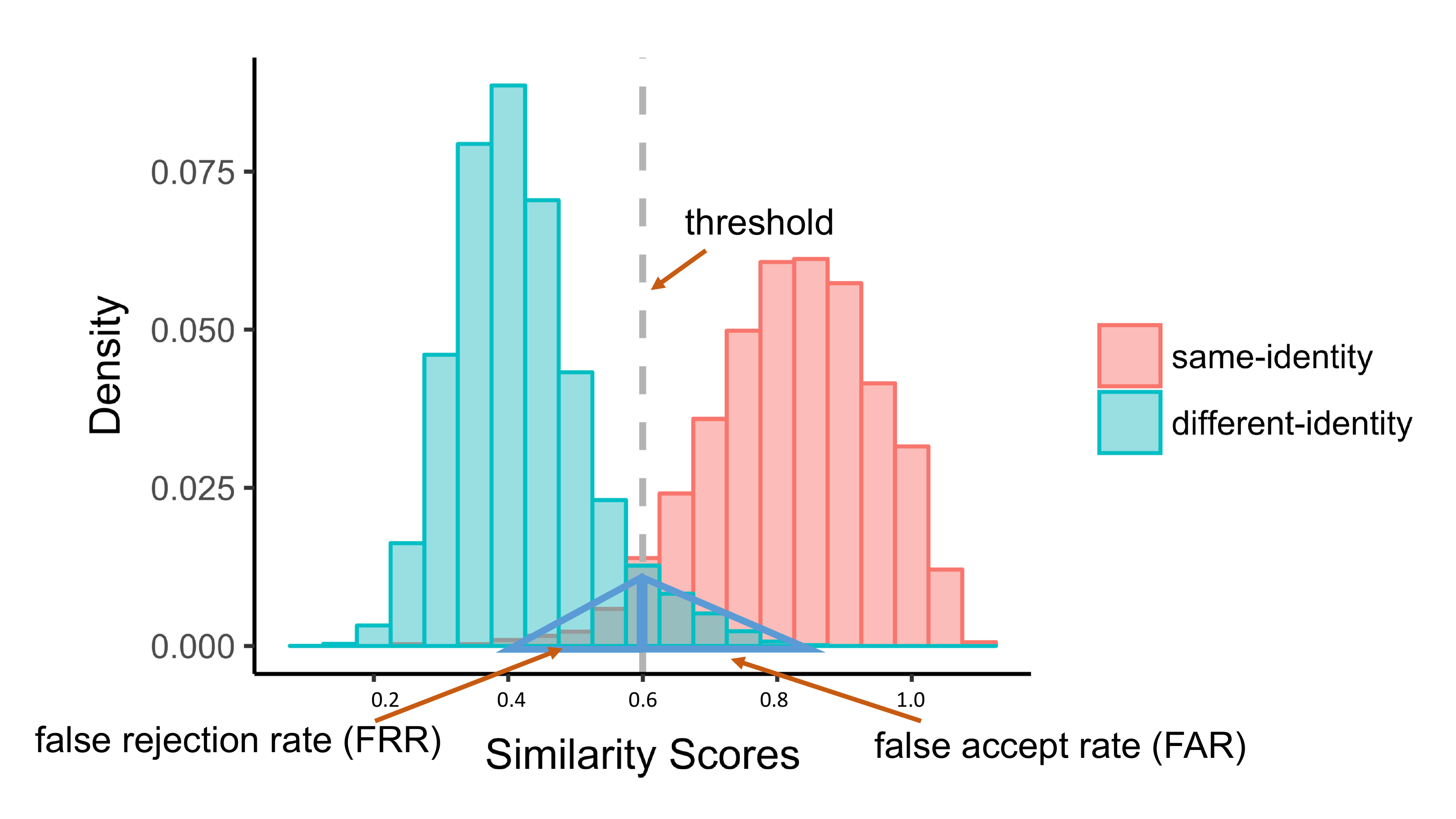}
\vskip -0.25cm
\caption{Signal Detection Model of Identification Accuracy. Similarity score histograms for same-identity image pairs (pink) and different-identity image pairs (teal) are shown. Decision thresholds (gray dotted line) specifies the similarity score cut-off for identification and determines false rejection and false accept rates.}
\label{SDT}
\vskip -0.45cm
\end{figure}

A common practice is to set a threshold that yields a very small proportion of false accepts (e.g., 1/1,000, 1/10,000 or less). In an application, the critical measure of accuracy then becomes the verification rate (VR) (same-identity pairs correctly judged as same-identity pairs) at the user-set threshold.  A threshold function shows FAR as a function of similarity score. As noted, threshold functions may differ by face race. We will discuss this issue in greater detail in the next section.

{\bf Lesson 3.} Algorithm accuracy measures can summarize the overall performance of a system (ROC and AUC) based on the underlying distributions. Or, they can summarize performance given both a user-set threshold {\it and} the underlying distributions. Consequently, identification can be measured with threshold-independent (ROC, AUC) or threshold-dependent measures (e.g., VR @ FA=0.001). Both types of measures serve a useful function, but they are not guaranteed to converge. As we shall see in Section 2, it is easy to understand how algorithms, which appear to perform at ceiling for two races of faces using an AUC (threshold-independent) measure, actually show small, but potentially meaningful, differences in VR at the very low FARs (threshold-dependent) required for scenarios. Therefore, AUC, as a summary measure of the full range of potential thresholds, can sometimes be a misleading indication of race bias. This is especially problematic when a machine is required to operate at extreme points in the ROC function. That said, at these extreme operating points performance differences may apply to a small number of cases (i.e., stimuli in the tails of the distribution).

\subsection{Factors underlying race bias}
\vskip -0.25cm
The underlying elements of race bias in algorithm performance are easily seen once we understand how the problem of face identification is formulated. The issues can be divided into data-driven and scenario-modeling issues.  

\subsubsection{Data-driven problems}
It is clear that the race bias in algorithm performance stems from the underlying same- and different-identity distributions. The dilemma is that the distribution parameters (e.g., means, standard deviations, skews) that characterize the population as a whole, may differ for particular demographic subgroups (race, age, gender). Differences among the subgroup distributions explain both everything and nothing about the underlying problem of bias. What we would like to know is {\it why} these distributions differ. Obvious possibilities include the following. 

First, it is possible that an algorithm produces representations that differ in quality (or uniqueness) for faces in different subgroups. Characteristics of the algorithm, such as the architecture or training data, may inadvertently produce biased representations. An algorithm with poor quality representations for particular subgroups would show race biases with any dataset. When the quality of the representation depends on adequate training with representative faces of a subset, we should expect bias. The racial composition of training datasets has been shown to effect the accuracy of previous generation algorithms (e.g., \cite{o1991simulating,klare2012face}). 

The problem for DCNNs may be more challenging, because they must be trained with enormous numbers of real-world (unconstrained) images. These images vary, not only across demographics, but also in quality and image factors (illumination, viewpoint, etc.). Therefore, it is difficult to isolate the impact of the racial composition of the training set on race bias. Because of the complexities of assessing the impact of training sets on DCNNs, there are currently no relevant results for these algorithms in the literature. Here, we concentrate on bias in datasets used to {\it test} DCNNs. These affect estimates of performance across race.

Second, demographic subgroups may be disproportionately represented in the test (i.e., measurement) distributions, potentially resulting in errors in the estimation of algorithm performance for certain demographic groups. For example, algorithms that operate on a specific population, should include test images representative of the individuals in the population. Also, some caution is warranted in assuming categorical structure for race with limited justification. Specifically, in biological terms, race is not categorical, and if we consider the issue from the more realistic perspective of mixed-race individuals, the problem is more complex.

Third, it is possible, that the quality of the training or test photographs differs across subgroups\cite{S_2019_CVPR_Workshops,krishnapriya2020issues,cook2019demographic}. For example, Krishnapriya et al. \cite{S_2019_CVPR_Workshops} found that image quality differences for the test faces explained some, but not all, of the variation in race bias. When only ICAO-compliant images were used to measure race bias, African American and Caucasian face accuracy improved across all four algorithms. The data on which race bias measurements are computed can contribute  to variable estimates of race bias in face recognition algorithms.

Fourth, it is possible that nested subgroups of faces have characteristics that amplify or hide bias effects. For example, race bias effects may be stronger for female faces than for male faces; or race bias may occur for challenging stimuli, but not for easier stimuli. We will see an example of this latter case in the experiment we present.  

{\bf Lesson 4.} Data-driven sources of race bias are defined objectively by differences in the underlying distributions of demographic sub-populations of faces. There are multiple reasons why these differences might occur. The underlying factors are not mutually exclusive. Therefore, in any given scenario, one or more data-driven anomalies might contribute to system bias.
 
\subsubsection{Scenario-modeling}
These are directly under the control of the ``user'' and include: (a) thresholds effects and (b) the formulation and control of the demographic homogeneity of the different-identity distribution. Beginning with thresholds, it is clear that a uniform threshold is not adequate or equitable when the underlying sub-population distributions differ. This is a concern with race bias. Previous studies have shown that the threshold needed to achieve a consistent FAR and/or VR varies across racial groups \cite{krishnapriya2020issues,S_2019_CVPR_Workshops,o2012demographic}. These differences underscore the importance of threshold shifts. Setting a single threshold for all racial groups may produce different estimates of FAR and VRs for different sub-groups of faces. This can be a source of race bias in the operation of algorithms.

For example, O'Toole et al. (2012), measured identification accuracy for Asian and Caucasian faces for two different stimulus sets using ROC curves and threshold functions \cite{o2012demographic}. ROC curves showed a slight advantage for Asian faces on the easier stimulus set, and an advantage for Caucasian faces on the more difficult stimulus set. However, the threshold functions, which show the relationship between similarity score and FAR, differed for the two datasets. For both datasets, there was a greater threshold shift for Asian faces than for Caucasian faces. This threshold difference indicates that achieving equitable FARs across both racial groups, would require setting a larger threshold for Asian faces compared to Caucasian faces.

Threshold differences have been demonstrated also for African American and Caucasian faces \cite{S_2019_CVPR_Workshops,krishnapriya2020issues}. As  noted, ROC curves showed that two out of four algorithms were less race biased for African American faces compared to Caucasian faces. However, despite this advantage, threshold shifts for African American faces differed uniformly for all four algorithms. For African American faces, all four algorithms needed a higher threshold to achieve a given FAR, and a lower threshold to achieve a given VR. These results provide additional evidence that uniform thresholds across race groups may produce different accuracy results for each race group. This study \cite{S_2019_CVPR_Workshops} also provides evidence that ROC curves can obscure information that threshold functions reveal. {See Lesson 3.}

A second scenario-modeling concern involves the demographic homogeneity of the different-identity distribution. Common practice is to measure algorithm accuracy with all possible different-identity pairs as the baseline. However, this practice can be problematic when different-identity pairs also differ demographically (e.g., by race, gender, age)\cite{o2012demographic}. 
If this is the case, the distribution for different-identity pairs will include demographic differences, in addition to identity differences. Clearly, the average similarity score computed for different-identity pairs that also differ in demographic group will be lower than the average similarity score for different-identity pairs within a homogeneous demographic. This demographic heterogeneity can shift the position of the different-identity distribution leftward, thereby artificially inflating algorithm verification performance. A solution to this problem is ``yoking". Yoking is defined as controlling the different-identity distribution so that all different-identity pairs are demographically comparable (e.g., same race, age, gender) \cite{o2012demographic}. This constraint produces a more homogeneous different-identity distribution (compared to including all possible different-identities in the analysis) and provides a more accurate assessment of accuracy and race bias.

Previous research demonstrates that yoking alters estimates of algorithm accuracy \cite{o2012demographic,o2017five,howard2019effect}. O'Toole et al. first demonstrated the effect of yoking on identification performance measures for Asian and Caucasian faces \cite{o2012demographic}. Different-identity distributions were demographically controlled in four groups: no control, race only, gender only, race and gender. The authors found that overall accuracy decreased with increased demographic control.

\textbf{Lesson 5.} Scenario-modeling problems are (partially) under the control of the researcher and can directly impact estimates of race bias. System-users should consider how thresholds and yoking controls affect each race of interest independently.

\section{Race Bias in Face Identification Algorithms}
\vskip -0.25cm
In this section, we present an experiment that examined race bias as a function of stimulus difficulty, using items previously calibrated for challenge-level (see Stimuli). Four face recognition algorithms (one pre-DCNN and three DCNNs) were tested on Caucasian and East Asian face pairs.  We also demonstrate the effects of yoking across race, and across race and gender, on estimates of identification accuracy across older and newer generations of DCNNs. This experiment also serves to highlight demographic bias issues that challenge algorithms in the context of the lessons we have learned from previous work.

\subsection{Methods}
\vskip -0.25cm
\subsubsection{Stimuli}
Images were taken from a condensed set of the Good, Bad, Ugly (GBU) Challenge dataset \cite{phillips2011introduction}. The  partitions of the GBU dataset provide us with the opportunity to compare race bias across algorithms and items (image pairs) that vary in difficulty. The GBU dataset contains images partitioned into three difficulty levels, referred to as: the Good, the Bad, and the Ugly. These difficulty partitions are based on the similarity scores of the top three performers in the Face Recognition Vendor Test (FRVT) 2006 \cite{Phillips:2009zl}. The full GBU set contains 1,085 images of 437 identities in each partition. The dataset includes frontal images that vary in environment (e.g., indoors and outdoors) and demographics (i.e., race, age, gender). The stimuli were captured using a Nikon D70 6-megapixel single-lens reflex camera.\footnote{The identification of any commercial product or trade name does not imply endorsement or recommendation by NIST.} 

All images were collected within a single academic year at the University of Notre Dame [see \cite{phillips2011introduction} for full details on how the GBU was compiled]. For the purposes of this study, only a subset of the full GBU dataset was selected. First, we restricted our analysis to images taken indoors, which limited the variation in illumination as a confounding factor. Second, given that East Asian and Caucasian faces represented the majority of images, only these two racial groups were selected for comparison. Although this reduced the race groups we were able to test, we chose the GBU dataset for testing race bias because (1) race in the GBU dataset is self-reported, (2) the conditions under which the images were taken are identical across and within race groups, and (3) age range is narrow, which limits other confounding demographics. For this reason, our analysis was performed on a condensed version of the GBU dataset (good: 385 identities; bad: 389 identities; ugly: 380 identities) (see Fig. 2 for a stimulus example).

\begin{figure}[h]
  \centering
  \includegraphics[width=\linewidth]{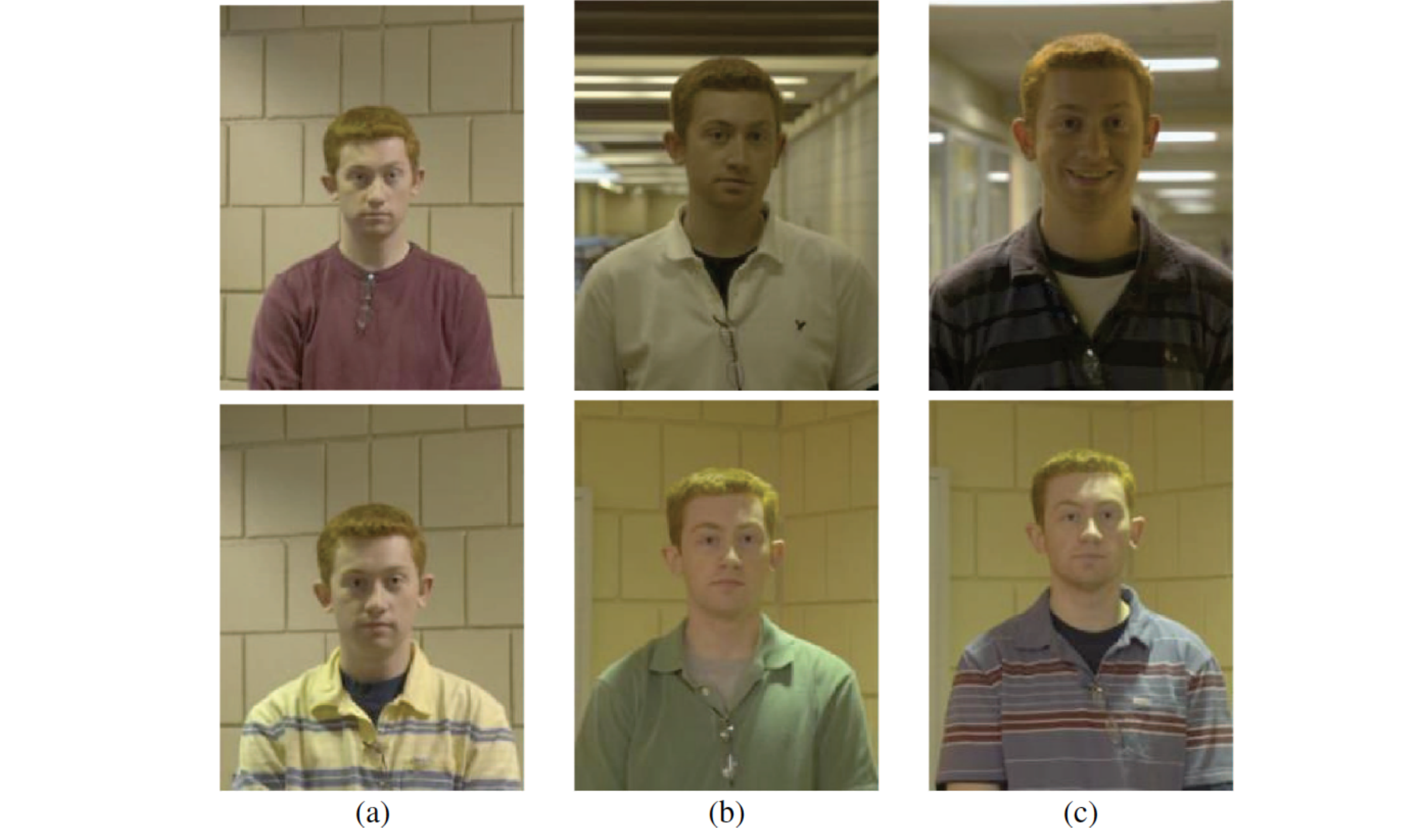}
  \caption{Example image pairs from the Good (a), the Bad (b), and the Ugly (c) difficulty partitions. All six images are the same identity \cite{phillips2011introduction}.}
  \label{Stimuli}
  \vskip -0.45cm
\end{figure}

\subsubsection{Algorithms}
Four algorithms were used to compare verification accuracy for East Asian and Caucasian faces: A2011  \cite{phillips2011introduction}, A2015 \cite{parkhi2015deep}, A2019 \cite{ranjan2019fast}, A2017b \cite{ranjan2017l2}. None of the algorithms were trained on face images collected at Notre Dame\footnote{The training datasets for A2011 are not known. For the DCNN algorithms the training datasets are described in the original papers.}. The A2011 algorithm is a fused algorithm of three top performing algorithms in the FRVT 2006 conducted by NIST. This fused algorithm pre-dates DCNNs, and thus A2011 is the oldest algorithm we tested. This algorithm was selected as a race-bias assessment of a pre-DCNN face recognition algorithms. Moreover, A2011 has been widely used in other comparisons \cite{phillips2017cross,phillips2014comparison}.  A2015 is a publicly available DCNN and is commonly as a benchmark for performance of DCNNs. A2015 is an older, but well established, DCNN that produces an output of 4,096-dimensional feature vectors. Previous studies have shown a race bias (greater verification accuracy for images of Caucasians compared to images of Blacks) for A2015 \cite{el2016face,S_2019_CVPR_Workshops}. The A2015 was selected because of its prominence in the literature as a baseline measure of algorithm accuracy \cite{el2016face,phillips2017cross,phillips2018face,S_2019_CVPR_Workshops}. A2017b, which produces 512-dimensional feature vector, is based on a ResNet-101 architecture and is trained on about 5.7 million images. Most recently, A2017b was found to be comparable in accuracy to that of forensic facial examiners \cite{phillips2018face}. Finally, A2019 is based on an Inception architecture that produces 512-dimensional feature vectors also, and is trained on close to one million images. To our knowledge, no study has previously examined how A2019 and A2017b perform on different racial groups. Both A2017b and A2019 were selected because they represent state-of-the-art algorithms with well-defined training data. This study provides the first direct comparison of race bias across a pre-DCNN algorithm, a previous generation DCNN, and 2 high-performing DCNN algorithms. All are published and available for scrutiny.

\subsection{Results}
\vskip -0.25cm
\subsubsection{Overall Accuracy}
We computed accuracy by collapsing across race and GBU distributions to derive a single ROC curve for each algorithm. ROC curves and AUC scores were calculated to measure face verification performance accuracy. ROC curves for all figures are plotted on a log-scale function to show performance at commonly set FAR = 0.0001 and 0.001 (gray dotted lines) and are race and gender yoked. Overall accuracy was best for the two newer algorithms, A2019 and A2017b, followed by A2011, and A2015 (Fig. 3).

\begin{figure}[h]
  \centering
  \includegraphics[width=\linewidth]{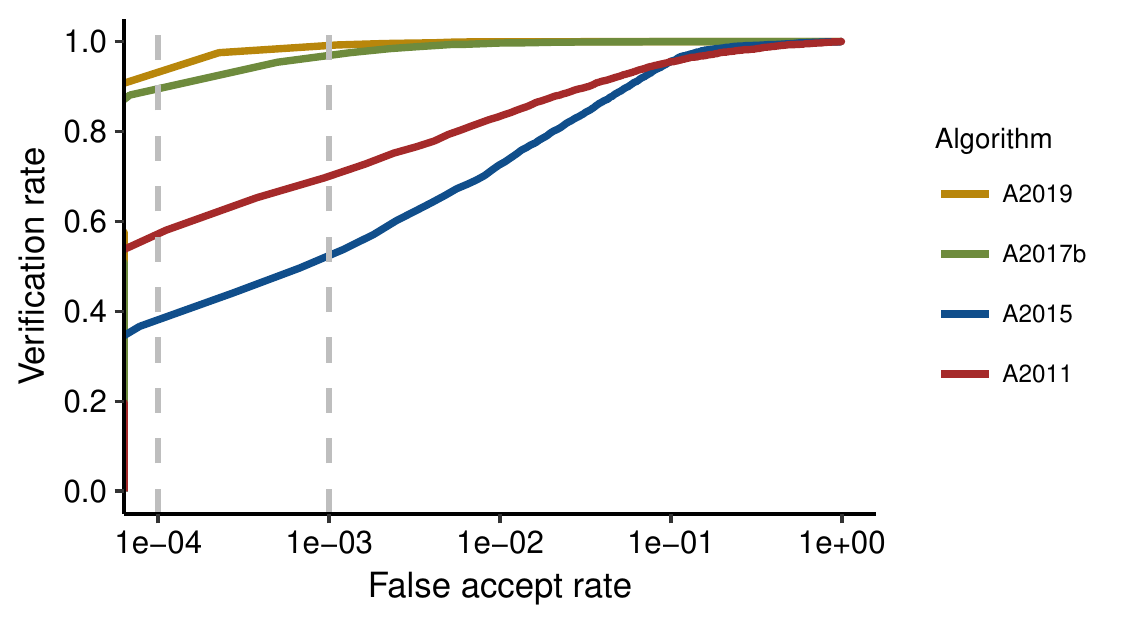}
  \caption{ROC curves for  A2019 (yellow), A2017b (green), A2015 (blue), A2011 (red).  A2019 and A2017b show near perfect performance, followed by A2011, and A2015.}
  \label{Overall Performance}
  \vskip -0.45cm
\end{figure}

\subsubsection{Yoking}

Similarly to overall accuracy, ROC curves were derived by collapsing across GBU distributions. The effects of yoking are easily seen on overall accuracy by comparing results with three yoking conditions. For the {\it no yoking} condition, all available different-identity pairs were considered, regardless of cross-race and cross-gender status of identities. For the {\it race-yoking}  condition, only same-race different-identity pairs were considered. For the {\it race and gender-yoking} condition, only same-race and same-gender different-identity pairs were considered. Fig. 4 shows that algorithms showed yoking effects in the predicted directions (accuracy decreased as yoking constrains increased), but that the magnitude of the effects varied. The effect of yoking on overall accuracy was most evident for both A2011 and A2015, and less so for A2017b and A2019. (See {\bf Lesson 5}).

\begin{figure}[h]
    \includegraphics[width=\linewidth]{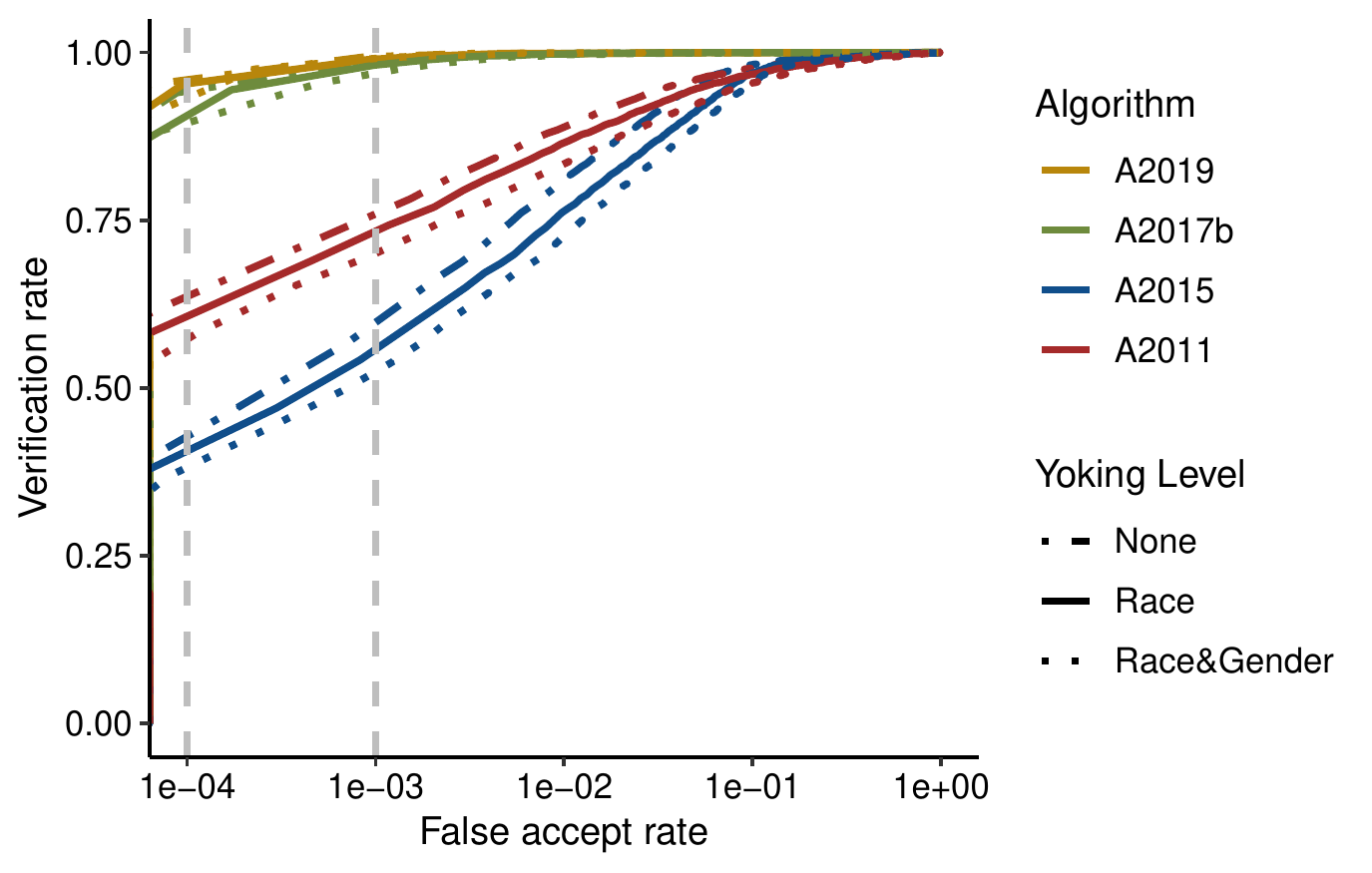}
    \vskip -0.25cm
  \caption{Yoking effects on accuracy for A2019 (yellow), 2017b (green), A2015 (blue), A2011 (red), for no yoking (dashed), race yoking (solid), and race and gender yoking (dotted).}
  \label{yoking}
  \vskip -0.50cm
\end{figure}

\subsection{Race Bias}
\vskip -0.30cm
\subsubsection{ROC Curves}

Verification accuracy on East Asian and Caucasian
faces was calculated. AUC results for algorithms A2015, A2019 and A2017b appear in Table~\ref{tab:Race-accuracy}. As can be seen, by the AUC measure, performance is near ceiling (AUC = 1.0) in all cases. By this account, all four algorithms show little to no race bias on overall accuracy for Caucasian and East Asian faces. Examination of the verification estimates at low FARs, however, tell a different story (Fig. 5). These reveal effects of race bias for all algorithms. For A2019 and A2017b, there was no or little bias for FARs larger than 0.001; however, for smaller FARs there was bias. These results demonstrate a case where overall accuracy (threshold-independent) and accuracy at a pre-determined threshold (threshold-dependent) may lead to different conclusions, despite the fact they are internally consistent (See {\bf Lesson 2} and {\bf Lesson 3}).

\begin{table}[h]
  \caption{AUC of algorithms A2011, A2015, A2017b, and A2019 on Caucasian faces and East Asian faces.}
  \label{tab:Race-accuracy}
  \centering
  \begin{tabular}{lcc}
    \toprule
     & \multicolumn{2}{c}{AUC}\\
     \cline{2-3}
    Algorithm&Caucasians&East Asians\\
    \midrule
    A2011 & 0.981614 & 0.977027\\
    A2015 & 0.990328 & 0.973814\\
    A2017b & 0.999670 & 0.999205\\
    A2019 & 0.999674 & 0.999886\\
  \bottomrule
\end{tabular}
\vskip -0.45cm
\end{table}

\begin{figure}[h]
  \centering
  \includegraphics[width=\linewidth]{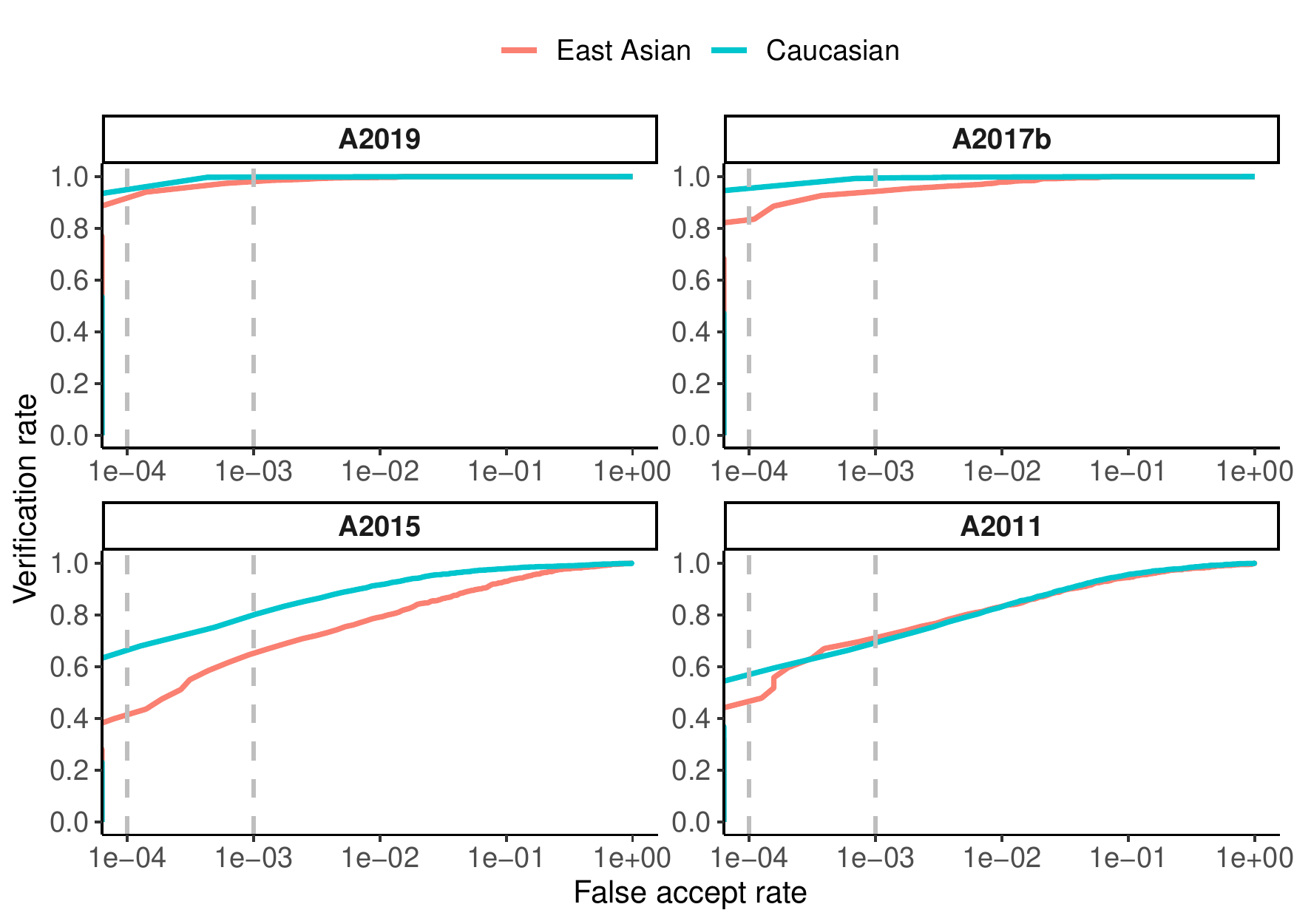}
  \caption{ROC curves for Caucasian (teal) and East Asian (orange) faces for A2019 (top-left), 2017b (top-right), A2015 (bottom-left), A2011 (bottom-right). 
  AUC measures showed race bias for A2015 only. However, ROC curves at low FARs (gray dotted lines) show that all algorithms perform more accurately for Caucasian faces than for East Asian faces.}
  \label{Race}
  \vskip -0.45cm
\end{figure}

\subsubsection{Thresholds}

We calculated FAR as a function of the different-identity similarity-scores for East Asian and Caucasian faces (see Fig. 6). For all four algorithms, the East Asian threshold function (orange) is always to the right of the Caucasian function (teal)\footnote{Because the similarity score scales differ across algorithms, direct comparisons for threshold shift magnitudes across algorithms is not possible.}. Because of this shift, a fixed threshold will yield a smaller FAR for Caucasian faces than for East Asian faces. To obtain the same FAR for both races will require separate thresholds for both races. (See {\bf Lesson 3} and {\bf Lesson 5}). 

\begin{figure}[!t]
  \centering
  \includegraphics[width=\linewidth]{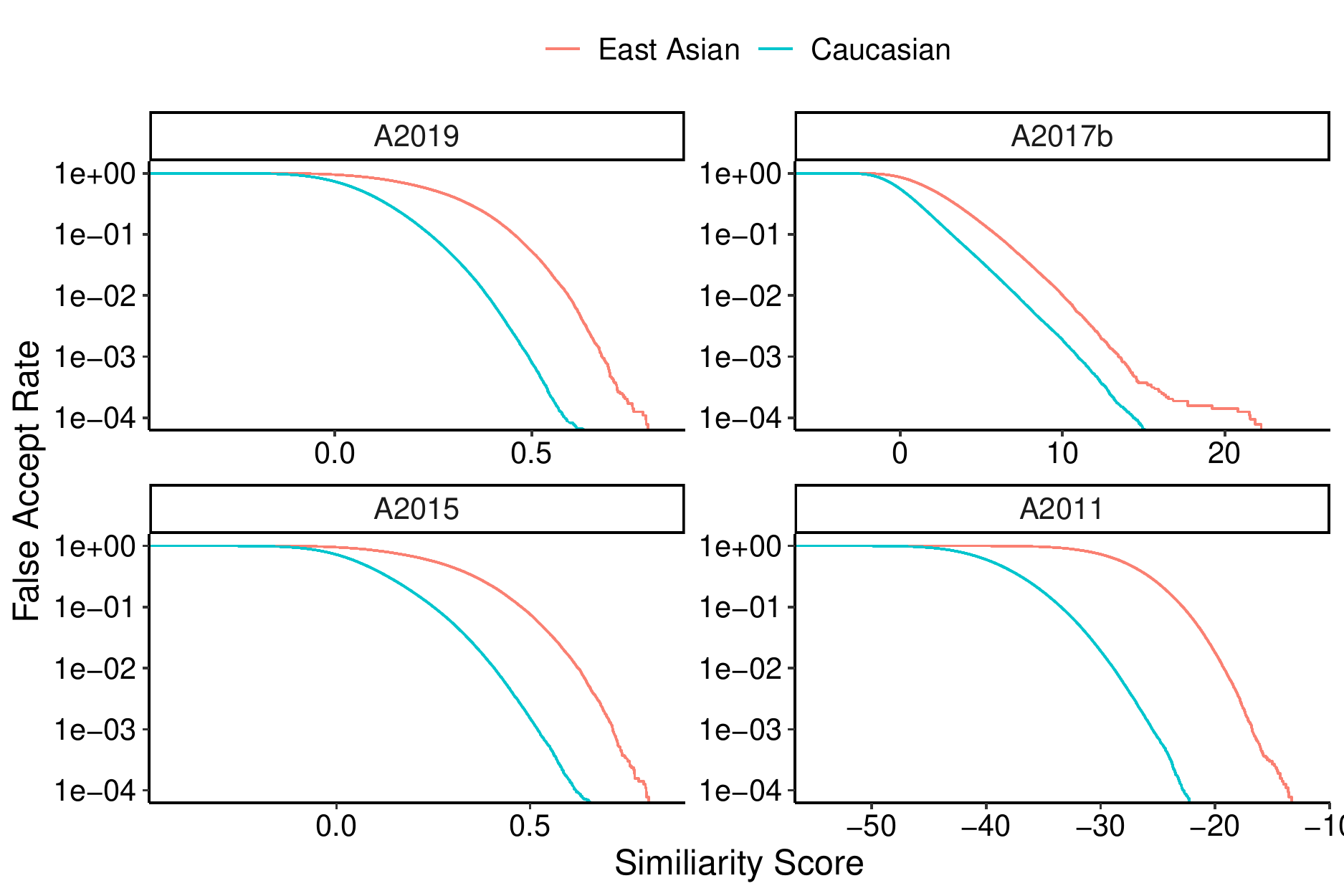}
  \caption{Threshold functions for Caucasian (teal), East Asian (orange) faces for  A2019 (top-left), A2017b (top-right), A2015 (bottom-left), A2011 (bottom-right). The plot shows a rightward shift for East Asian faces (orange) relative to Caucasian (teal) faces indicating that equivalent FARs require a greater threshold for East Asian faces.}
  \label{Threshold}
\end{figure}

\subsection{Item Difficulty}
\vskip -0.25cm
Performance on each GBU partition was calculated (Fig. 7). A2019 and A2017b were more accurate than A2015 and A2011 across all three partitions. A2015 had the lowest verification accuracy for the Good and Bad images, whereas A2011 had the lowest accuracy for Ugly. These data are comparable to previous research that shows the trade-off between A2015 and A2011 in the Bad and Ugly partitions \cite{phillips2017cross}. Accuracy for the Ugly partition was the lowest across all three difficulty groups. 

\begin{figure}[!t]
  \centering
  \includegraphics[width=\linewidth]{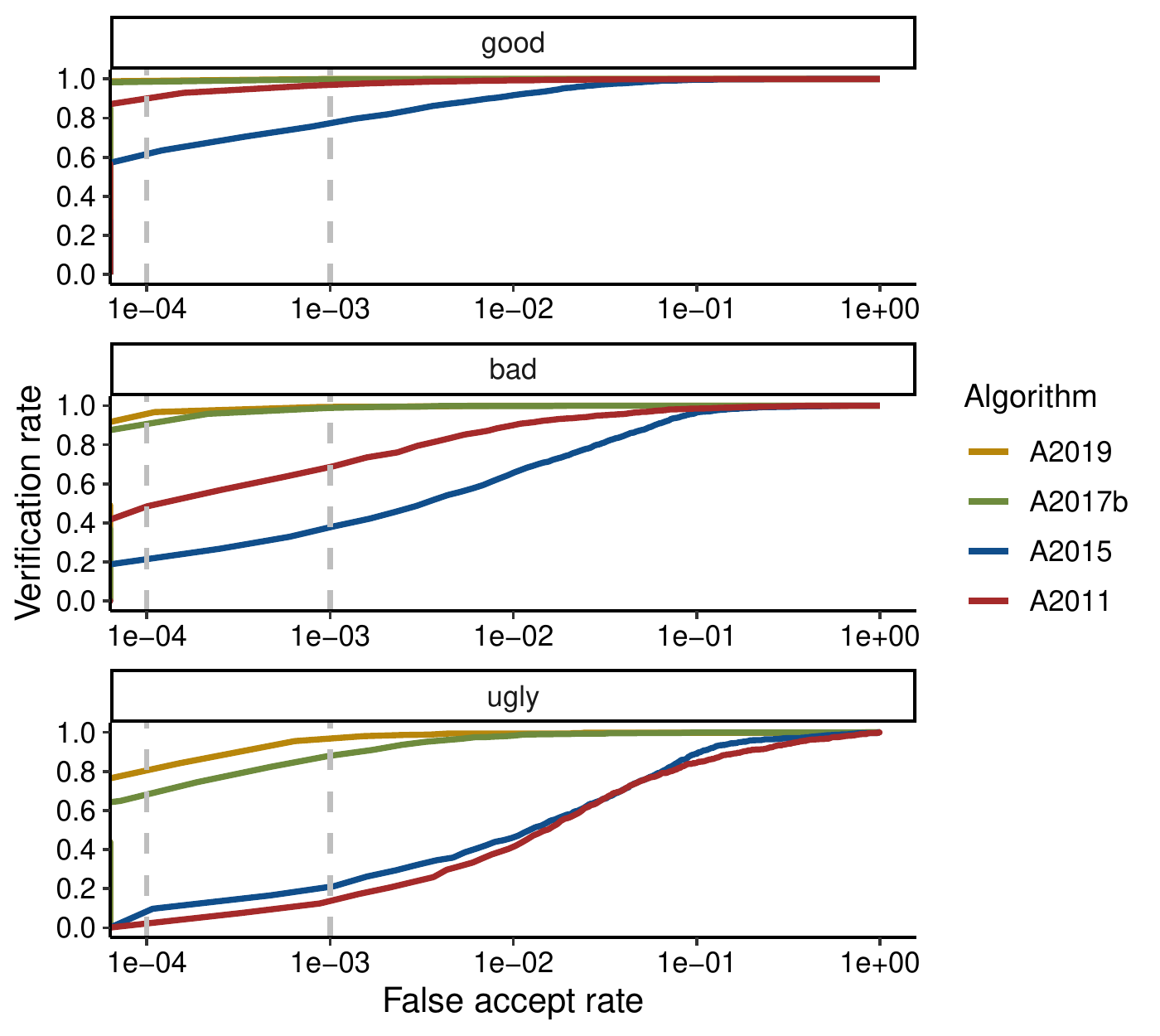}
  \caption{ROC curves for Item Difficulty. Accuracy for A2019 (yellow), 2017b (green), A2015 (blue), A2011 (red) on Good (top panel), Bad (middle panel), Ugly (bottom panel). A2019 and A2017b were the most accurate across all three partitions.}
  \label{OverallGBU}
  \vskip -0.45cm
\end{figure}

\subsubsection{Race Bias as a Function of Item Difficulty}
The GBU dataset allows us to examine a novel problem for DCNNs, race bias as a function of well-defined item difficulty. We computed ROC curves for East Asian and Caucasian faces across the three difficulty levels of GBU (Fig. 8). Accuracy for the Good partition is nearly perfect for both East Asian and Caucasian faces. A2015, and to a lesser extent A2011, showed a race bias in favor of Caucasian faces at FAR = 0.0001. For the Bad partition, A2017b and A2019 again showed nearly no race bias, whereas A2011 and A2015 showed greater verification accuracy Caucasian faces at FAR = 0.0001. For the Ugly partition, no algorithm achieved perfect performance for either race. All algorithms, except for A2011, showed greater accuracy for Caucasian faces compared to East Asian faces.
(See {\bf Lesson 4}).

\begin{figure*}
    \includegraphics[width=\textwidth]{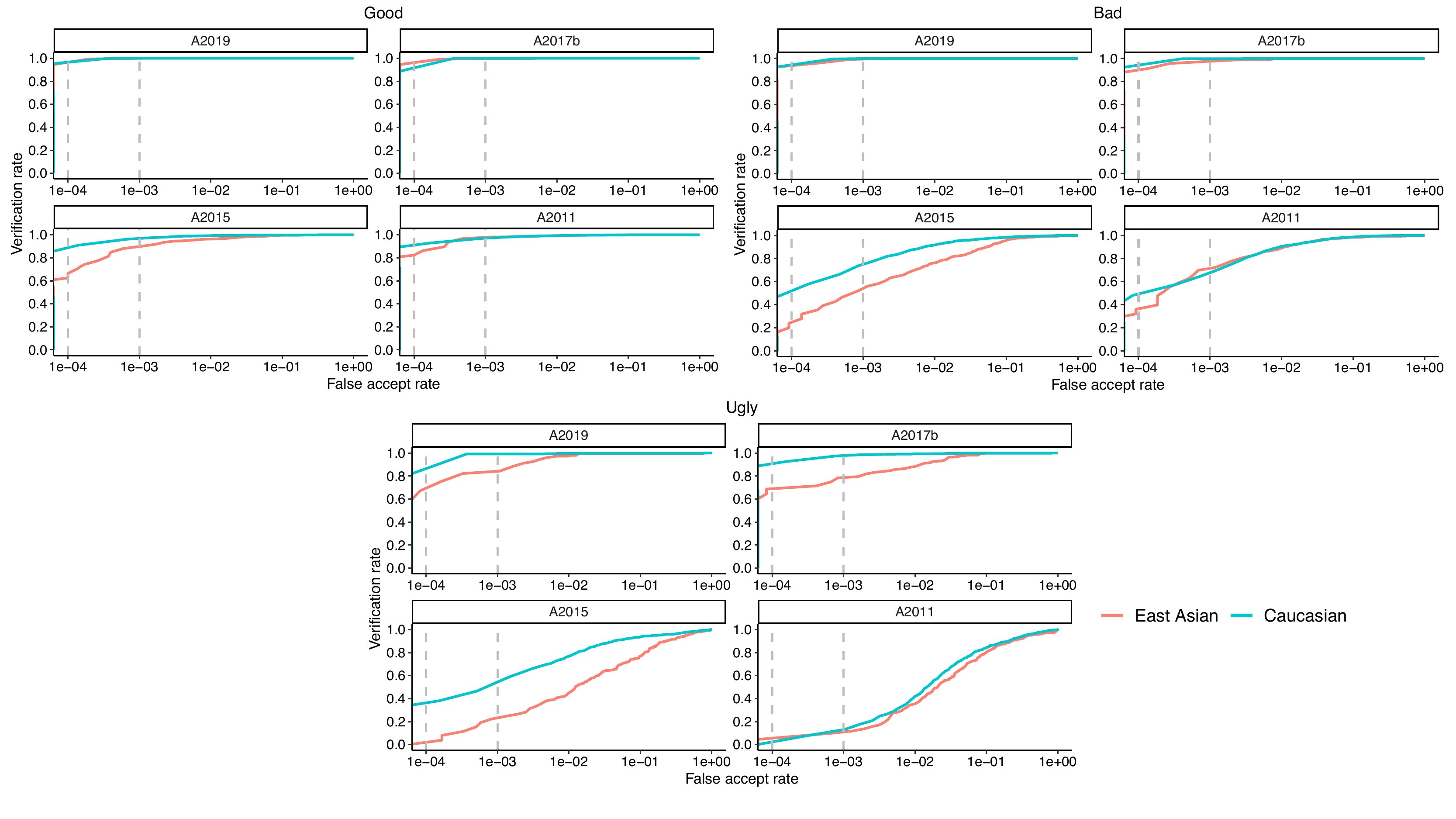}
  \caption{East Asian and Caucasian ROC Curves for GBU partitions (A2019, top-left; A2017b, top-right; A2015, bottom-left;  A2011, bottom-right). Top panel: Good and Bad partitions (left and right) show nearly perfect accuracy for A2019 and A2017b for all faces. A2015 shows greater accuracy for Caucasian faces. 
  Bottom panel: Ugly partition, shows some degree of race bias for all algorithms except A2011.}
  \label{GBURace}
  \vskip -0.45cm
\end{figure*}

\section{Discussion}
\vskip -0.25cm
We reviewed and analyzed literature published over the last 50 years from humans and algorithms on race bias in face recognition. Five lessons in measuring race bias for algorithms emerged in our review of past work. These five lessons inform our understanding of the data-driven and scenario-modeling factors that impact race bias in face recognition algorithms. These factors are applied to novel empirical data, which we collected on race bias as a function of item pair difficulty for three recent DCNN algorithms and one pre-DCNN algorithm.  

To begin, we review data-driven and scenario-based factors that impact race bias. The sources of bias that a researcher must consider include the following:
 
\begin{itemize}[leftmargin=*]
    \item {\bf Data-driven factors:}
	\begin{itemize}
	    \item {\it sub-population distributions}: the population statistics for demographic groups
	    \item {\it algorithm}: the quality of the algorithm's representations across demographic groups
	   \item {\it representative images}: the subgroup's representation of the population of interest
	    \item {\it imaging conditions:}  the imaging conditions directly affect the difficulty of comparing images 
         \end{itemize}	
    \item {\bf Scenario-modeling:}
	\begin{itemize}
		\item {\it threshold}: appropriate selection of threshold for a desired FAR for each racial group, independently
	   \item {\it demographic-pairing}: modeling the homogeneity of the different-identity distribution 
         \end{itemize}
          
 \end{itemize}

This is the first study to consider the full range of factors that impact race bias in face recognition algorithms. Attention to race bias factors in past work has often been piecemeal. For example, common focus has been on the role of training data, and although this is an important factor, there remains a broad scope of other factors to consider. Data-driven factors underlie real differences in an algorithm's capacity to recognize faces of different races. Scenario-based factors are part of the measurement process and affect estimates of algorithm bias. The complexity of these factors, and their potential to interact, makes a general assessment of bias for face recognition algorithms unfeasible. Consequently, race bias must be measured for each particular scenario, algorithm, race, and dataset. 

The empirical data presented here provides two additional key contributions to the burgeoning literature on race bias. First, an updated analysis of race bias across three generations of face recognition algorithms (pre-DCNN, older DCNN, and two newer DCNNs) provides insight on the evolution of race bias. We find that with each new generation of algorithms, accuracy improves for both race groups. However, results from threshold-dependent measures suggest that accuracy differences across race groups at specific points of interest remain problematic. Second, we provide novel findings on the impact of item difficulty on race bias accuracy. Specifically, as item challenge level increased demographic differences were magnified.

One  limitation in our study was the use of only two racial groups in our analysis. As noted, the GBU dataset was selected because of the excellent control it provides of factors other than race. This photometric and demographic control makes it ideal for studying race bias, but somewhat limits the ecological validity. Although the analysis was limited to two groups, the lessons learned and methodological considerations apply across all race groups. Moreover, the focus on only two races allowed us to explore a wide range of possible factors that may contribute to bias.

This brings us back to the question that motivated this work. Where are we on measuring race bias? Our findings point to strong improvements across racial groups and concomitant declines in race bias overall. However, these gains may vary as a function of item difficulty--a factor that has not been considered previously in assessing race bias for algorithms. With the rise of DCNNs, the complexity of the bias problem increases, due to the need for extremely large training sets and the large number of parameters that may impact performance on subsets of faces. Clearly, race bias in face recognition algorithms is a critical problem that remains unsolved. Although promising approaches are on the horizon (cf., \cite{gong2019debface}), these methods need to be tested more thoroughly. In many cases, they are still constrained to work on specific challenges that may, or may not, generalize to other types of problems. Until a technical solution to the problem of bias is found, algorithms need to be tested individually and thoroughly for performance across racial groups. Our holistic assessment is integral to understanding that the solution to the race bias problem is not simple or straightforward. The intricacy of this issue is underscored by the multitude of underlying factors that impact race bias. From an applied perspective, each application needs to measure the bias that accounts for the algorithms, data-driven factors and scenario conditions. In addition, bias needs to be continuously monitored, because of changes in the data characteristics, demographic shifts, and algorithm updates.

\section{Conclusion}
\vskip -0.25cm
The five lessons we present provide a starting point for developing a principled understanding of how race and demographic bias should be assessed for face recognition algorithms. We also considered how image difficulty impacts these estimates of race bias and how this has changed with the evolution of face recognition algorithms. At this point in time, advances in the field require simultaneous attention to all potential sources of race bias in face recognition algorithms. Both data-driven and scenario-modeling factors and their interactions can impact race bias in face recognition algorithms. This holistic assessment provides a realistic and informed starting point for future studies in this area.


\ifCLASSOPTIONcompsoc
  \section*{Acknowledgments}\vskip -0.25cm

\else
  \section*{Acknowledgment}\vskip -0.25cm

\fi

Funding: Supported by the Intelligence Advanced Research Projects
	Activity (IARPA). This research is based upon work supported by the Office of the Director of National Intelligence (ODNI), Intelligence Advanced Research Projects Activity (IARPA), via IARPA R\&D Contract No. 2014-14071600012 and 2019-022600002. The views and conclusions contained herein are those of the authors and should not be interpreted as necessarily representing the official policies or endorsements, either expressed or implied, of the ODNI, IARPA, or the U.S. Government. The U.S. Government is authorized to reproduce and distribute reprints for Governmental purposes notwithstanding any copyright annotation thereon. Support for J.C. was provided by National Eye Institute Grant	1R01EY029692-01 to AOT.



%
\bibliographystyle{IEEEtran}
\bibliography{references}

\end{document}